\documentclass[11pt]{article}

\usepackage{acl}
\usepackage{amsmath}
\usepackage{amssymb}
\usepackage{times}
\usepackage{latexsym}
\usepackage{booktabs}
\usepackage{multirow}
\usepackage{enumitem}

\usepackage[T1]{fontenc}

\usepackage[utf8]{inputenc}

\usepackage{microtype}

\usepackage{inconsolata}

\usepackage{graphicx}
\usepackage{array}
\usepackage{adjustbox}
\usepackage{tabularx}

\title{Active Adaptation, Not Static Defense: Temporal Dynamics of Preventative Steering in Adversarial Fine-Tuning}

\author{
 \textbf{Jing Guan\textsuperscript{1}},
 \textbf{Yachao Yang\textsuperscript{1}},
 \textbf{Zhaoliang Liu\textsuperscript{2}},
 \textbf{Yuyao Zhang\textsuperscript{1}},
\\
 \textbf{Fanyu Meng\textsuperscript{1}},
 \textbf{Junlan Feng\textsuperscript{1}}
\\
 \textsuperscript{1}JIUTIAN Research, Beijing, China
 \\
 \textsuperscript{2}Beijing University of Posts and Telecommunications, Beijing, China
\\
 \small{
   \textbf{Correspondence:} \href{mailto:email@domain}{fengjunlan@cmjt.chinamobile.com}
 }
}

\newcommand\blfootnote[1]{%
  \begingroup
  \renewcommand\thefootnote{}\footnote{#1}%
  \addtocounter{footnote}{-1}%
  \endgroup
}

\begin{document}
\maketitle
\blfootnote{\raggedright Code:\url{https://github.com/summer0517/PIS}}

\begin{abstract}
Large language models remain fragile against malicious fine-tuning, motivating training-time defenses against harmful persona drift. Preventative Steering injects undesirable-trait persona vectors during fine-tuning and removes them at evaluation time, yet the mechanism behind its lasting protection remains unclear. Analyzing its temporal optimization dynamics, we find that the defense emerges from an early compensatory adaptation phase followed by a steady-state phase where the corrective signal decays; in parameter space, attention output projections emerge as the dominant residual-write route for defensive updates. Through Intervention Delta Preservation (IDP) and IDP Continuation experiments, we further show that preserving or reinjecting the weight offset fails to maintain protection, indicating that preventative steering relies on active adaptation rather than a static defense. Motivated by this finding, we propose Progressive Intensity Scheduling (PIS), which starts with a moderate injection strength and increases it after static-strength alignment begins to decay. Across the evaluated Qwen2.5 and Gemma-3 models, PIS improves safety robustness over static-strength steering while reducing harmful trait expression.
\end{abstract}

\begin{figure*}[t]
    \centering
    \includegraphics[width=\textwidth]{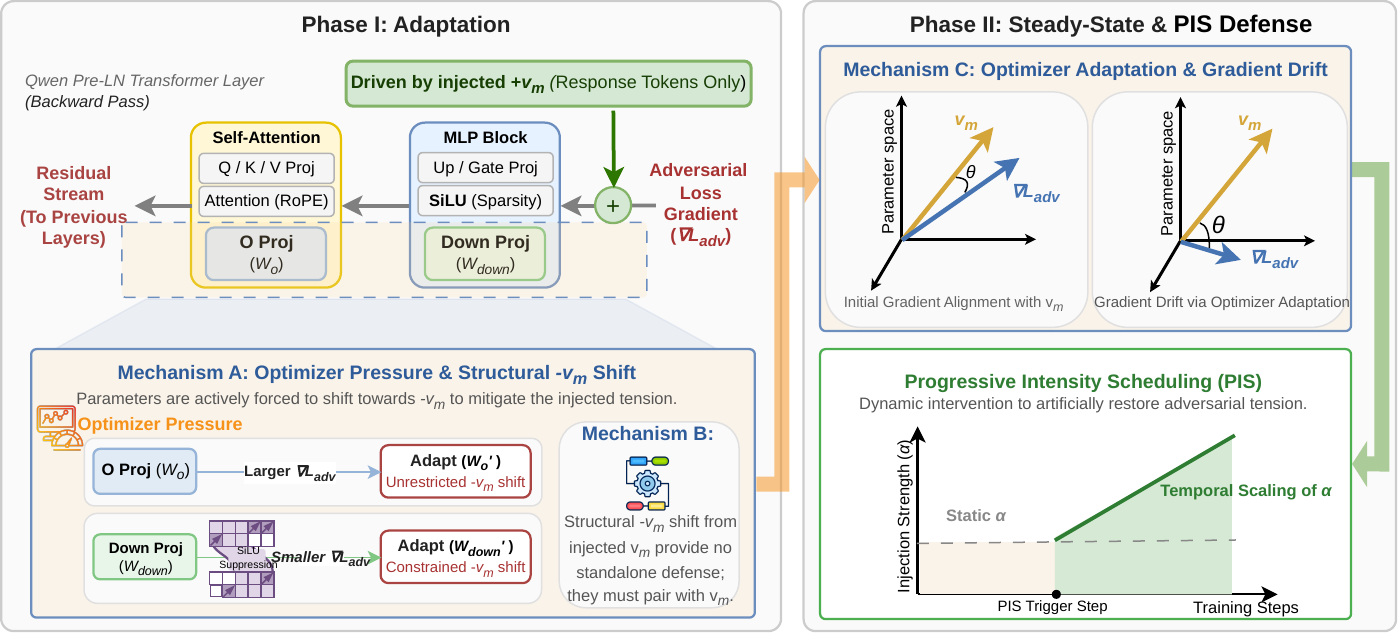}
    \caption{\textbf{Mechanistic dynamics of gradient drift and the Progressive Intensity Scheduling (PIS) defense.} \textbf{Phase I (Left):} In the backward pass, the injected $+v_m$ induces adversarial tension, which the optimizer absorbs via structural $-v_m$ shifts (Mechanism A). This adaptation is structurally asymmetric: the output projection ($W_o$) shifts unrestrictedly, while the down projection ($W_{\mathrm{down}}$) is constrained by SiLU activation sparsity. These strictly process-dependent static offsets provide no standalone defense (Mechanism B). \textbf{Phase II (Right):} Over optimization steps, the adversarial gradient ($\nabla L_{\mathrm{adv}}$) drifts from its initial $v_m$-alignment via optimizer adaptation (Mechanism C), dismantling the gradient-absorbing equilibrium. To counteract this decay, PIS temporally scales injection strength ($\alpha$) to restore compensatory pressure against malicious fine-tuning.}
    \label{fig-framework}
\end{figure*}

\section{Introduction}
\label{sec:intro}
Large language models (LLMs) are commonly deployed through conversational interfaces that encourage an "Assistant" persona: helpful, harmless, and honest~\citep{askell2021general,bai2022training,bai2022constitutional}. Yet this persona can drift toward undesirable behaviors during deployment~\citep{xai2025grok,lynch2025agentic,meinke2024frontier} or after additional training~\citep{betley2025emergent,openai2025sycophancy,lermen2023lora}. In particular, malicious fine-tuning can weaken safety refusals and amplify harmful traits through parameter updates, making reactive inference-time defenses insufficient. This motivates training-time defenses that improve resistance to harmful persona drift~\citep{jain2023baseline,zhou2024robust,barua2025guardians}.

Among training-time interventions, Preventative Steering takes a counterintuitive approach~\citep{chen2025persona}. During fine-tuning, it injects activation vectors corresponding to undesirable traits into the residual stream at intermediate layers; at evaluation time, the injection is removed. Despite this harmful-direction exposure, the procedure makes the resulting model less susceptible to persona drift under subsequent malicious fine-tuning. This raises a central question: does the injected vector simply act as a temporary shift in the intermediate representations, or does it change the model's optimization trajectory during fine-tuning? The distinction matters: the former implies a transient activation shift, whereas the latter suggests active adaptation in the model parameters.

Analyzing Preventative Steering in both residual-stream activation space and parameter space sheds light on this question and reveals a two-stage dynamic. Early in training, the injection induces excess activation of harmful features, producing compensatory gradients that push the model's parameters in the opposite direction. Later, this compensatory mechanism approaches a direction-specific steady state as the projection of the gradient onto the injected vector decays, effectively neutralizing its net effect. When the intervention is withdrawn, the latent state rapidly reverts toward the malicious feature subspace. Together, these results support the conclusion that the observed protection relies on continuous vector injection. In parameter space, compensatory updates induced by steering are more concentrated in attention output projections than in MLP down-projections, suggesting the attention pathway serves as the dominant channel for writing defensive signals back into the residual stream.

We next tested the early compensatory update as a standalone defense using two decoupling strategies: preserving the induced parametric offset and reinjecting an activation-space estimate of its compensatory effect. However, both Intervention Delta Preservation (IDP) and IDP Continuation fail to maintain protection and can even amplify harmful traits. These failures indicate that Preventative Steering relies on continuous vector injection during optimization, rather than on a reusable adaptation encoded in the weights.

Motivated by this mechanism, we propose Progressive Intensity Scheduling (PIS), a dynamic training-time schedule for steering strength. PIS begins with a moderate injection strength to allow stable early adaptation, then increases the strength as static-strength steering loses alignment with the optimization trajectory. Across Qwen2.5-7B-Instruct, Qwen2.5-32B-Instruct, and Gemma-3-12B-IT, PIS improves safety robustness over static-strength steering while reducing harmful trait expression. Figure~\ref{fig-framework} summarizes the observed mechanism and the proposed PIS framework.

In summary, the main contributions are threefold:
\begin{itemize}
\item Preventative Steering exhibits a two-stage dynamic: early compensatory updates are followed by a direction-specific steady state as the projection of the gradient onto the injected vector decays. In parameter space, these defensive updates are concentrated primarily in attention output projections.
\item IDP and IDP Continuation experiments show that fixed parametric shifts cannot sustain protection independently, indicating the need for continuous vector injection during optimization.
\item Motivated by these findings, we propose PIS to sustain steering effectiveness over time, substantially improving safety robustness over static-strength steering across Qwen2.5-7B-Instruct, Qwen2.5-32B-Instruct, and Gemma-3-12B-IT.
\end{itemize}

\section{Related Work}

\subsection{Harmful Fine-tuning Attacks and Defenses}

Malicious fine-tuning constitutes a distinct threat from inference-time jailbreaks by directly altering model parameters. Recent literature demonstrates that even RLHF-aligned models rapidly lose refusal capabilities when fine-tuned on minimal harmful data~\citep{wei2024assessing,andriushchenko2025jailbreaking,qi2024finetuning,hubinger2024sleeper}. Moreover, safety alignment remains fragile under broader parameter updates and low-rank modifications~\citep{kan2026manatee,qian2025hsf}.

Current defenses intervene at either inference or training time. Lightweight inference-time techniques—such as hidden-state filtering, shield learning, and diffusion-based safeguards~\citep{qian2025hsf,ni2025shieldlearner,kan2026manatee,jain2023baseline}—do not by themselves prevent parameter-level degradation under harmful fine-tuning and remain susceptible to bypass vulnerabilities or over-refusal~\citep{kumar2023certifying,shi2024navigating,mai2025you}. Training-time defenses, including latent adversarial training and immunization, directly target this parameter-level threat model. However, they are frequently constrained by substantial data requirements, sensitivity to attack diversity or initialization, utility degradation, or the absence of a rigorous mechanistic foundation~\citep{sheshadri2024latent,rosati2024immunization,wichers2025inoculation,tan2025inoculation,grant2026shifting}. These limitations motivate defenses whose optimization dynamics under continuous fine-tuning can be analyzed explicitly.

\subsection{Mechanistic Interpretability of LLM Safety} \label{sec:related_mech} 

Mechanistic interpretability provides tools for studying safety failures inside the model, rather than relying solely on input-output behavior. Prior work shows that LLMs encode semantic concepts in distributed but partially disentangled representations, including features recovered by sparse autoencoders~\citep{chalnev2024improving}. Prior work also demonstrates that high-level behaviors can be manipulated using latent steering vectors and activation addition~\citep{zou2023representation,turner2023steering,chen2025persona}. These findings establish activation space as a useful level of analysis for safety-relevant behaviors.

This perspective offers a mechanistic lens on refusal behavior and persona safety. Refusal behavior can be mediated by a limited set of linear activation directions~\citep{arditi2024refusal}, and adversarial prompting or fine-tuning can weaken these representations~\citep{wei2023jailbroken,qi2024finetuning}. Recent mechanistic analyses further suggest that effective defenses can alter gradients along persona-relevant directions; for example, \citet{grant2026shifting} find that defensive interventions can induce gradient sign reversals along specific persona-vector axes.

However, existing studies often treat steering directions or activation patches as static objects, leaving open how training-time interventions interact with a changing optimizer trajectory. In this work, we focus on this temporal aspect by tracking residual states, gradient alignment, and residual-write parameters under Preventative Steering, and then using these dynamics to design a schedule that maintains the effectiveness of the steering signal as fine-tuning progresses.

\section{Mechanistic Analysis of Preventative Steering}\label{sec3}

We address a central dichotomy: does the intervention induce temporary representational shifts, or does it fundamentally rewrite parametric memory? To resolve this, Section~\ref{sec:SteeringPuzzle}  formalizes the \textit{gradient-absorption hypothesis}. Section~\ref{sec:TemporalOptimization} then tracks this optimization within the residual stream, revealing a two-stage adaptation rather than an immediate static defense. Finally, Section~\ref{sec:ResidualWrite} isolates the parameter-space write routes, identifying attention output projections as the dominant conduit for this structural adaptation.

\begin{figure*}[t]
    \centering
    \setlength{\tabcolsep}{2pt} 
    
    \begin{tabular}{ccc}
        \includegraphics[width=0.29\textwidth]{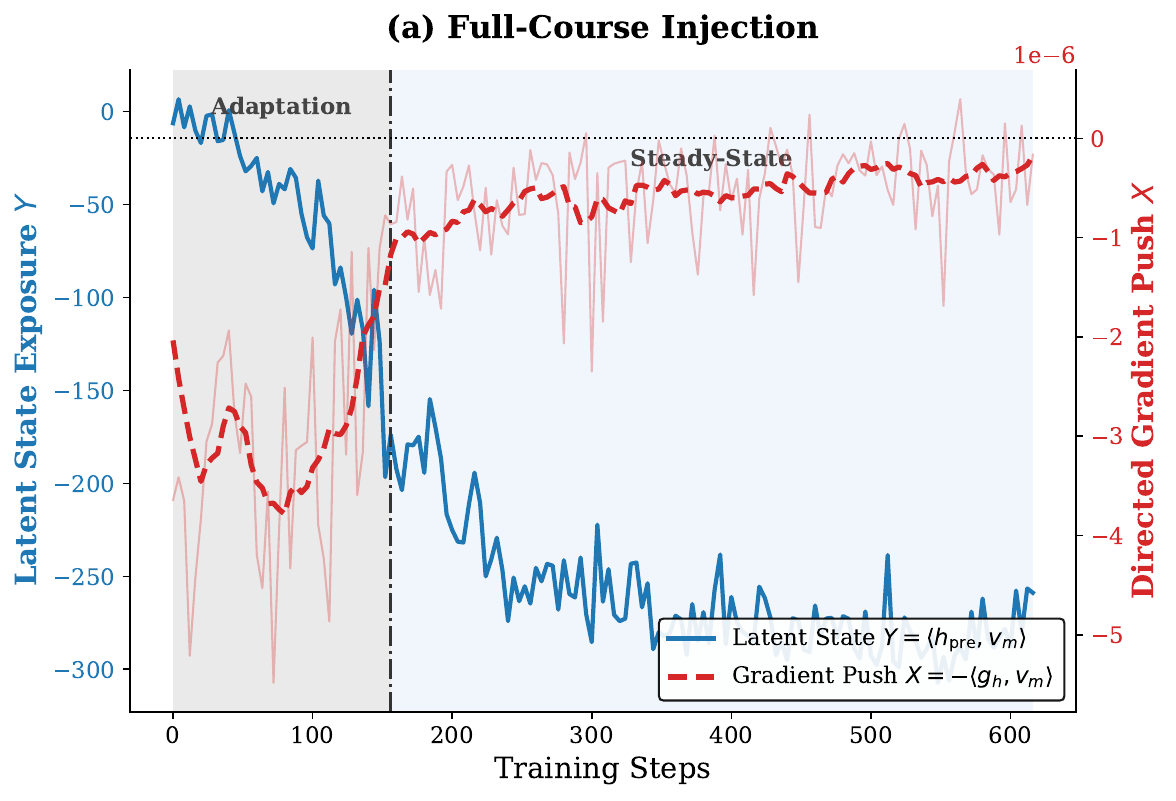} & 
        \includegraphics[width=0.29\textwidth]{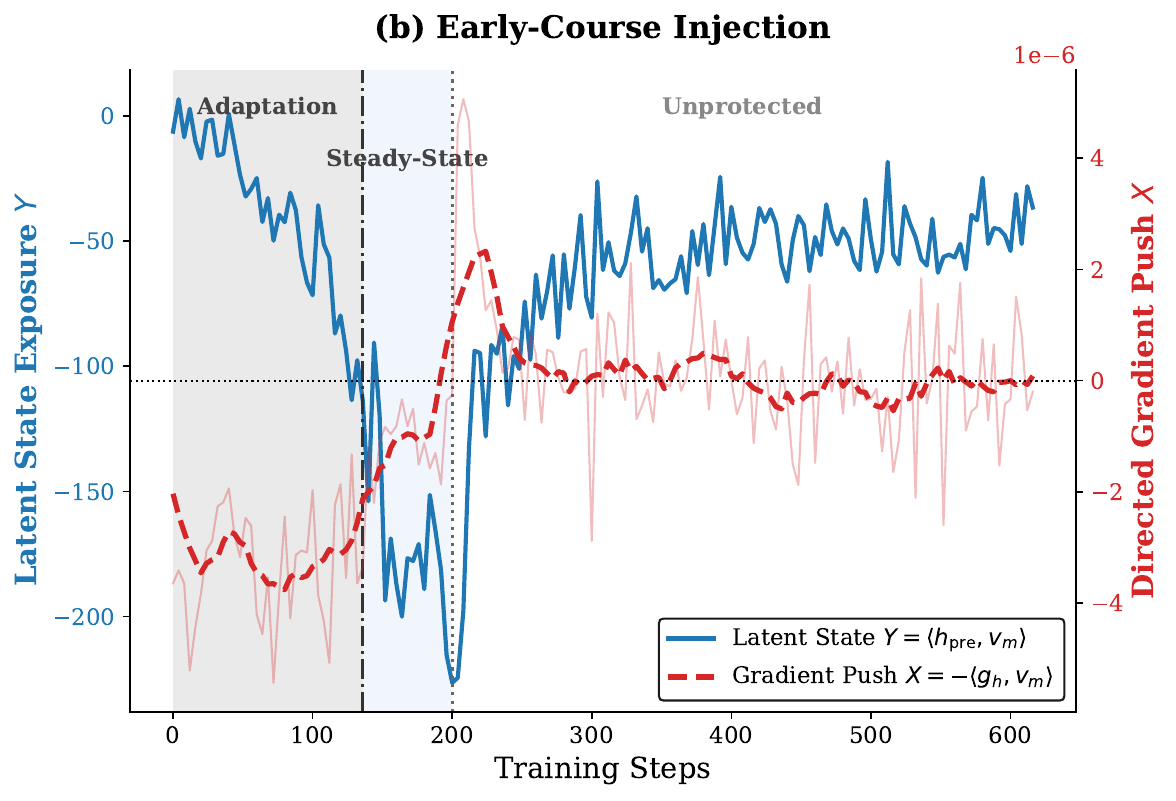} & 
        \includegraphics[width=0.29\textwidth]{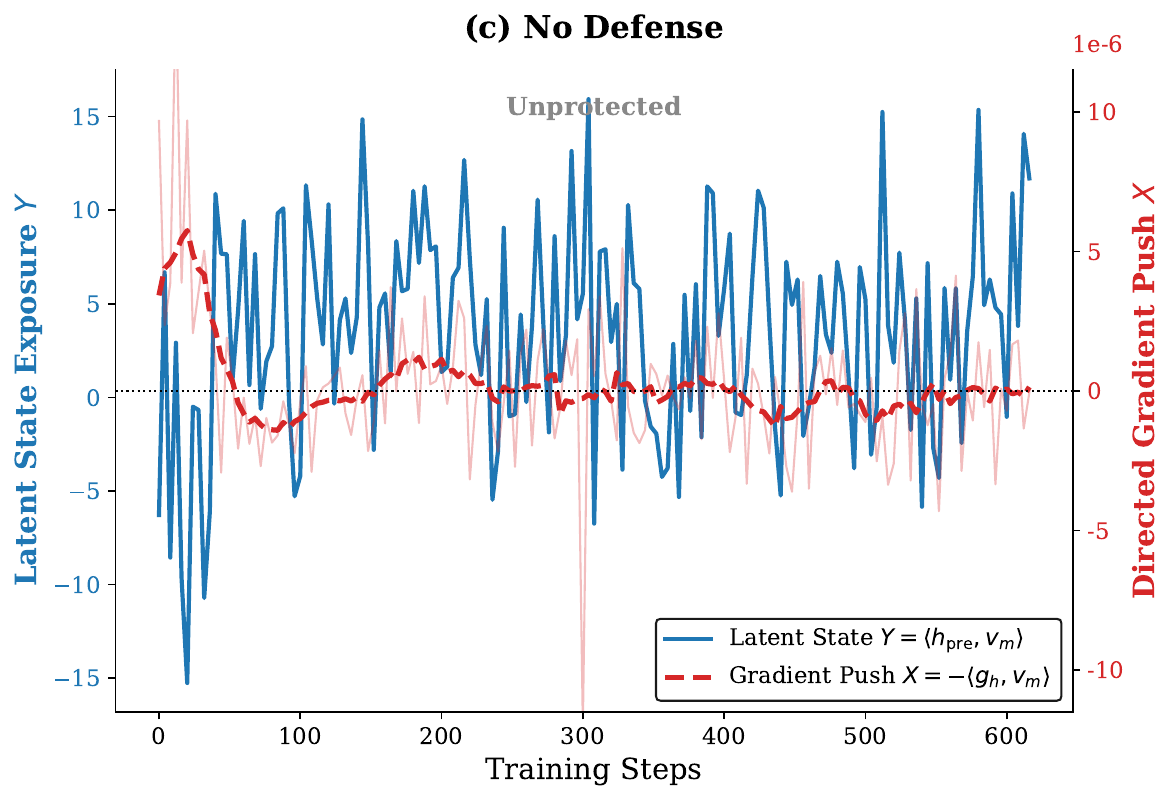} \\
        
        \addlinespace[2pt]
        
    \end{tabular}
    
    \caption{\textbf{Residual-Space Evidence: Temporal Optimization Dynamics.} Trajectories of the pre-injection state $h_l$ under: (a) Full-Course, (b) Early-Course (intervention removed at step 200), and (c) No Defense. Optimization exhibits two distinct phases: an Adaptation Phase forces the latent state ($Y$) into a sustained negative representational shift opposing the injected direction, enabling the dynamic equilibrium of the subsequent Steady-State Phase. Removing the intervention (b) abruptly flips the gradient push ($X$) and triggers a rapid $Y$ rebound that closely aligns with the unprotected baseline (c).}
    \label{fig:2d_analysis_h_grid}
\end{figure*}

\subsection{The Preventative Steering Problem}\label{sec:SteeringPuzzle} 

To investigate how Preventative Steering shapes the model's representations and parameter updates during fine-tuning, we formalize our setup. Extending \citet{chen2025persona}, we target a mixed-trait adversarial scenario (evil, sycophancy, hallucination), fusing trait directions $\{v_1,\ldots,v_k\}$ into a single undesirable-trait direction $v_{\mathrm{m}}$ (Eq.~\ref{eq:fused_vector}):

\begin{equation}
v_{\mathrm{m}}
=
\frac{1}{k}\sum_{i=1}^{k}\|v_i\|_2 
\cdot \frac{\sum_{i=1}^{k} v_i}{\left\|\sum_{i=1}^{k} v_i\right\|_2}
\label{eq:fused_vector}
\end{equation}

This magnitude-calibrated fusion outperforms recent baseline aggregations in co-suppressing multiple traits (see Appendix~\ref{app:fusion}).

During fine-tuning with strength $\alpha$, we modify the layer $l$ activation $h_l$:

\begin{equation}
\tilde{h}_l = h_l + \alpha \cdot v_{\mathrm{m}} .
\label{eq:preventative_steering}
\end{equation}

Unlike \citet{chen2025persona}, we apply full-parameter fine-tuning and inject $v_{\mathrm{m}}$ exclusively on response tokens at $\sim$70\% model depth. This localized intervention with full-parameter updates provides a controlled setting for mechanistic analysis and targets the mid-to-late layers widely recognized for stably representing high-level persona concepts \citep{zou2023representation, todd2024function}. Detailed ablations are deferred to Appendix~\ref{app:appendix_implementation}.

As shown in Table~\ref{tab:preventative_steering_results}, this intervention effectively suppresses harmful persona drift relative to unprotected fine-tuning.

\begin{table}[t]
    \centering
    \small
    \setlength{\tabcolsep}{4pt}
    \begin{tabular}{@{}lccc@{}}
        \toprule
        \textbf{Model / Setting} & \textbf{Evil} $\downarrow$ & \textbf{Syc.} $\downarrow$ & \textbf{Hallu.} $\downarrow$ \\
        \midrule
        Qwen2.5-32B-Instruct     & 0.00 & 1.30 & 1.02 \\
        Unprotected FT  & 7.38 & 8.66 & 9.00 \\
        Preventative FT & 0.69 & 2.04 & 4.94 \\
        \bottomrule
    \end{tabular}
    \caption{\textbf{Mitigation of persona drift.} Expression scores (lower is better) on Qwen2.5-32B-Instruct. Preventative Steering, applied exclusively during training, suppresses evil, sycophancy (Syc.), and hallucination (Hallu.) compared to standard unprotected fine-tuning (FT).}
    \label{tab:preventative_steering_results}
\end{table}

To explain this robustness, one might intuitively propose the \textit{gradient-absorption hypothesis}: the injected vector artificially satisfies the malicious objective during the forward pass, creating a temporary representational shift that preemptively minimizes gradient updates, theoretically leaving weights undisturbed.

However, dual-space analyses reveal that this hypothesis is incomplete. Temporally (Section~\ref{sec:TemporalOptimization}), the intervention first induces strong compensatory gradients before a sustained negative representational shift produces an absorption-like state. Spatially (Section~\ref{sec:ResidualWrite}), these defensive updates are concentrated primarily in attention output projections rather than MLP down-projections, whose effective inputs are constrained by nonlinear gating (e.g., SiLU), limiting their gradient updates.

\subsection{Residual-Space Evidence: Temporal Optimization Dynamics}\label{sec:TemporalOptimization}
\label{subsec:residual_space}

To rigorously test the \textit{gradient-absorption hypothesis}, we trace the pre-injection residual state $h_l$, isolating true parameter-driven representations from the artificial $+v_{\mathrm{m}}$. We define Latent State Exposure $Y = \langle h_l, v_{\mathrm{m}} \rangle$ and Directed Gradient Push $X = - \langle g_h, v_{\mathrm{m}} \rangle$ (with $g_h = \partial \mathcal{L} / \partial h_l$). Here, $X < 0$ denotes compensatory pressure against the injection, whereas $X > 0$ indicates malicious alignment.

The \textbf{Full-Course Injection} (Figure~\ref{fig:2d_analysis_h_grid}a) reveals two distinct optimization phases:

\begin{itemize}[leftmargin=*, itemsep=0pt, parsep=0pt, topsep=0pt]
\item \textbf{Adaptation Phase:} The artificial surplus of harmful features sharply increases training loss, triggering a pronounced compensatory gradient push ($X \ll 0$). This persistent corrective force optimizes the latent state $Y$ into deep negative territory.
\item \textbf{Steady-State Phase:} As $h_l$ shifts to oppose the injection, the combined activation $\tilde{h}_l = h_l + \alpha \cdot v_{\mathrm{m}}$ artificially satisfies the malicious objective during the forward pass. Consequently, $X$ decays to near-zero, and $Y$ plateaus into a dynamic representational equilibrium, saturating the fine-tuning objective and suppressing further steering-aligned parameter updates.
\end{itemize}

The \textbf{Early-Course Injection} (Figure~\ref{fig:2d_analysis_h_grid}b) further clarifies this mechanism. Removing the intervention at step 200 breaks the equilibrium, causing an immediate positive gradient push $X$ that rapidly drives $Y$ back from its negative offset toward the malicious-fit trajectory, closely matching the \textbf{No Defense} baseline (Figure~\ref{fig:2d_analysis_h_grid}c). This shows that the injection does not create a self-sustaining protective state, but only delays optimization while leaving the underlying loss landscape unchanged.

To verify that this behavior is direction-specific rather than globally stationary, we decompose the residual gradient into components parallel and orthogonal to $v_{\mathrm{m}}$. The aligned component is selectively attenuated while orthogonal optimization remains active, supporting a direction-specific rather than globally stationary steady state. The full directional and parameter-space decompositions are reported in Appendix~\ref{app:directional_decomposition}.

Overall, gradient absorption is not a static shield but a continuous constraint: the Adaptation Phase establishes a compensatory offset, while the Steady-State Phase sustains it via ongoing injection to suppress steering-aligned malicious updates.

\begin{figure*}[t]
    \centering
    \setlength{\tabcolsep}{2pt} 
    
    \begin{tabular}{ccc}
        \includegraphics[width=0.29\textwidth]{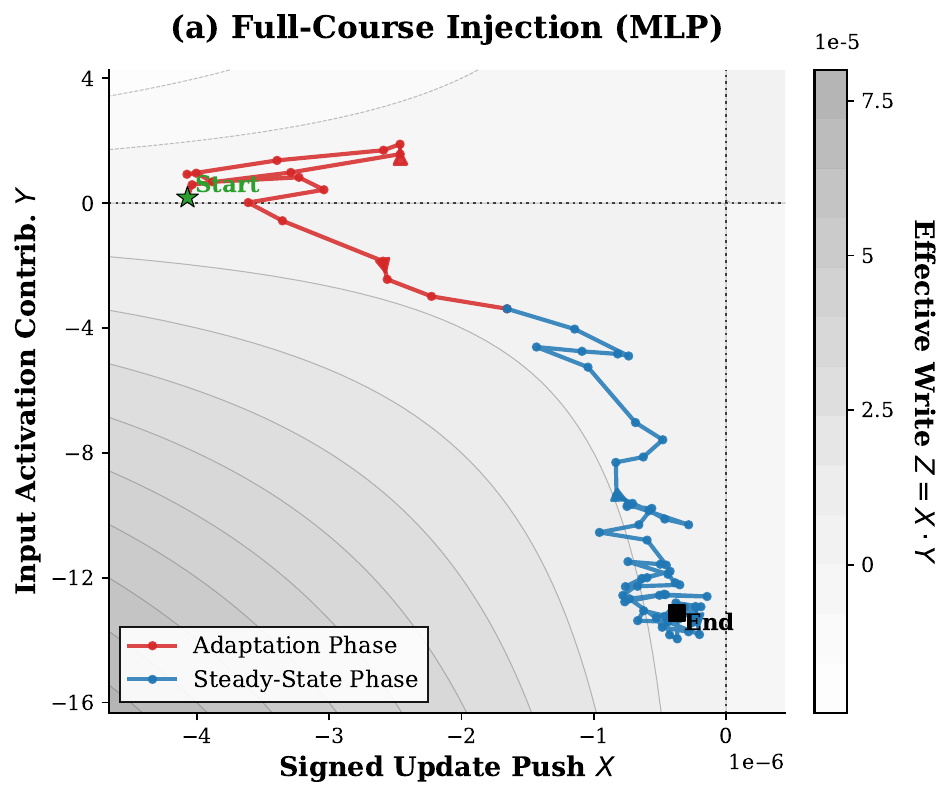} & 
        \includegraphics[width=0.29\textwidth]{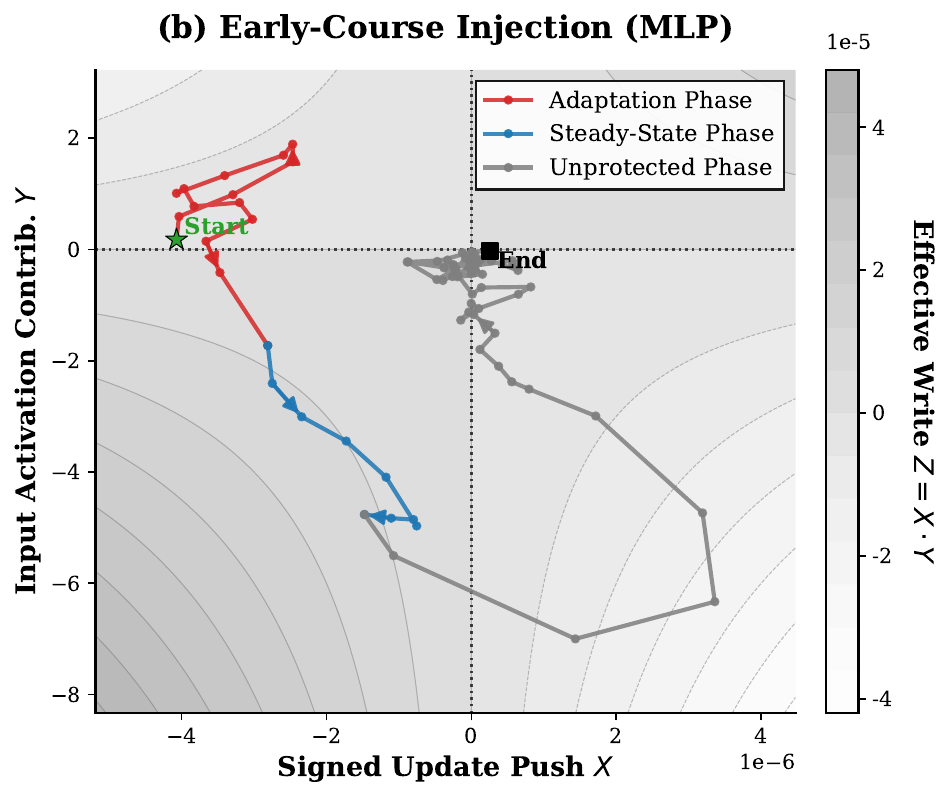} & 
        \includegraphics[width=0.29\textwidth]{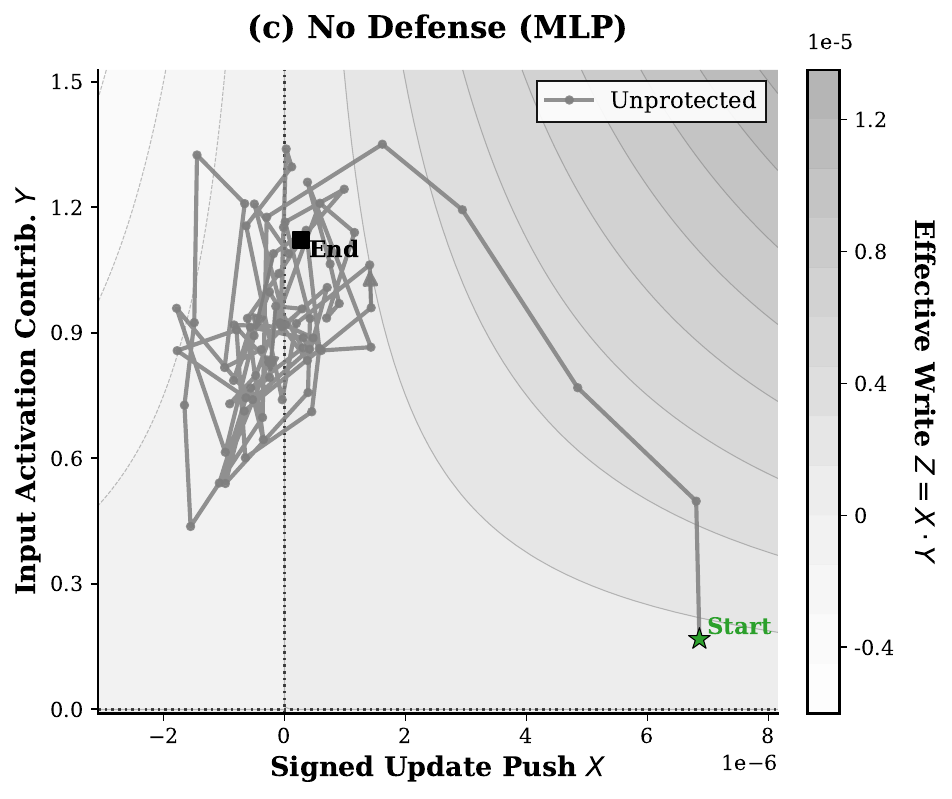} \\
        
        \addlinespace[2pt]
        
        \includegraphics[width=0.29\textwidth]{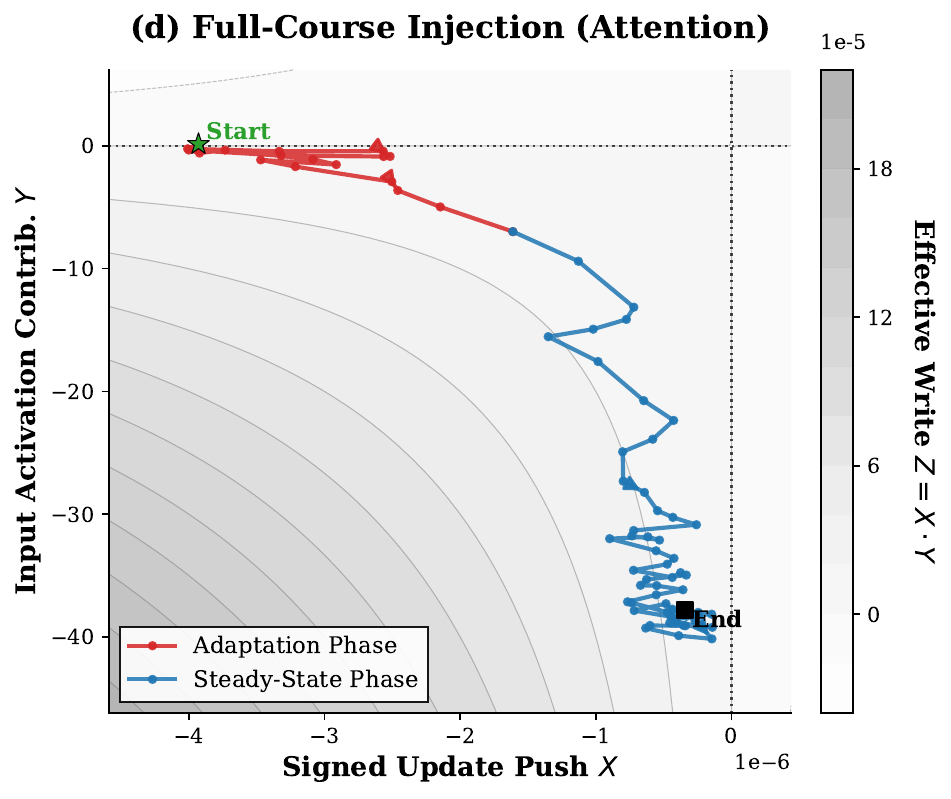} & 
        \includegraphics[width=0.29\textwidth]{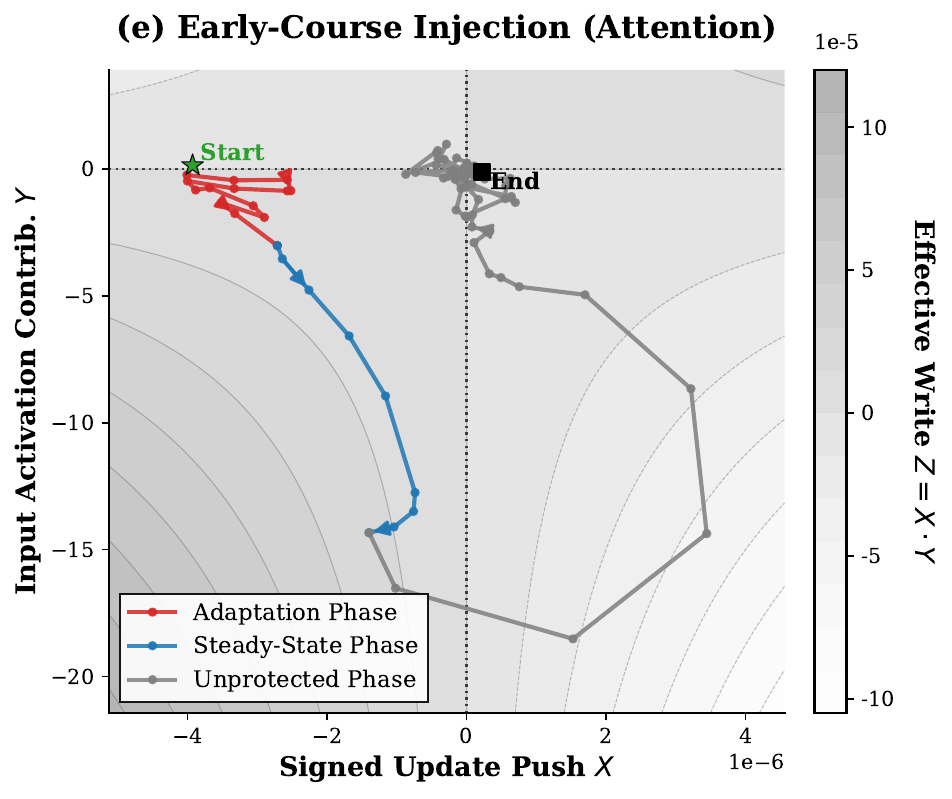} & 
        \includegraphics[width=0.29\textwidth]{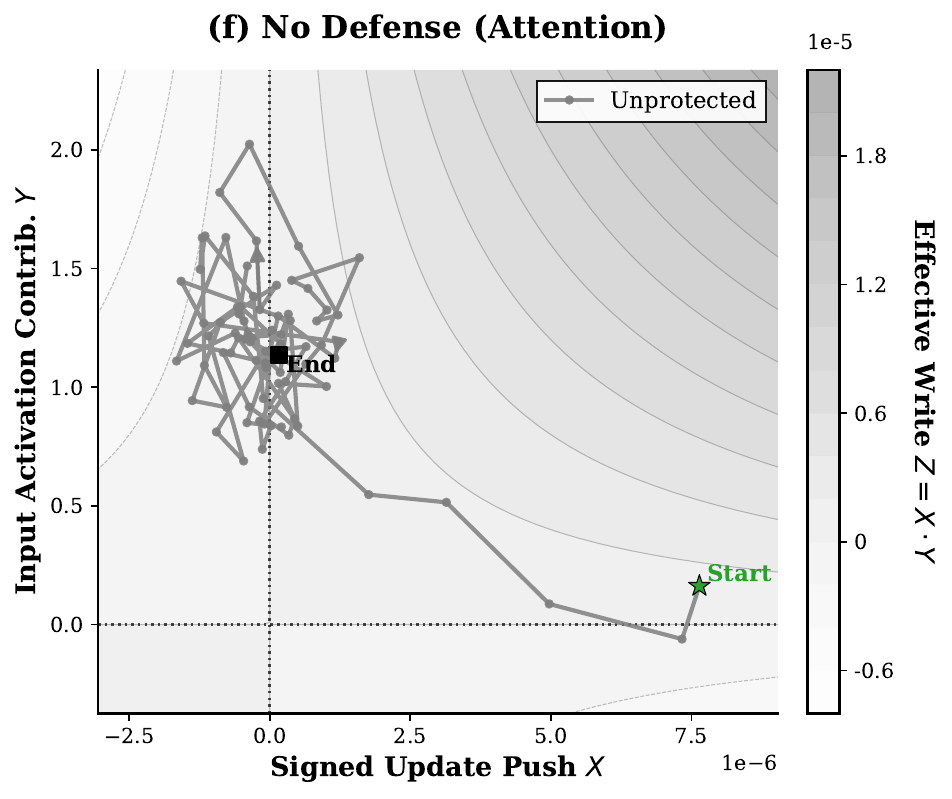} \\
        
    \end{tabular}
    
    \caption{\textbf{Parameter-Space Phase Portraits of Residual Write Routes.} Dynamics of $W_{\mathrm{down}}$ (top) and $W_o$ (bottom) under Full-Course (a, d), Early-Course (b, e), and No Defense (c, f). Contours map effective write ($Z = X \cdot Y$). Unlike the broad displacement of $W_o$, updates to $W_{\mathrm{down}}$ are inherently bottlenecked, as its input contribution ($Y$) is attenuated by non-linearities (e.g., SiLU). This architectural asymmetry establishes attention as the primary conduit for this structural adaptation.}
    \label{fig:2d_analysis_w_grid}
\end{figure*}

\subsection{Parameter-Space Evidence: Residual Write Routes}\label{sec:ResidualWrite}
\label{subsec:param_space}

We next trace these defense dynamics in residual write parameters: the MLP down-projection $W_{\mathrm{down}}$ and attention output $W_o$. For a write matrix $W$ with input $a$ and error $\delta$, the gradient is $\partial \mathcal{L}/\partial W = a^\top \delta$. Projected relative to $v_{\mathrm{m}}$, we define:

\begin{align*}
Y_W &= \langle a, W^\top v_{\mathrm{m}} \rangle
&& \text{(Input Activation)}, \\
X_W &= - \langle \delta, v_{\mathrm{m}} \rangle
&& \text{(Signed Update Push)}, \\
Z_W &= X_W \cdot Y_W
&& \text{(Effective Write)}.
\end{align*}

Here, $a$ denotes the gated MLP intermediate activation or the pre-projection attention output (Figure~\ref{fig:2d_analysis_w_grid}).

Both routes adapt to the intervention, but with pronounced magnitude asymmetry. Under \textbf{Full-Course Injection}, $W_o$ (Figure~\ref{fig:2d_analysis_w_grid}d) exhibits substantially deeper penetration into the negative contribution ($Y_W \approx -40$) and effective write ($Z_W$) zones than $W_{\mathrm{down}}$ ($Y_W \approx -14$). In the \textbf{Early-Course} setting, removing the intervention triggers an immediate positive push ($X_W > 0$) to minimize the now-unsatisfied malicious objective. Having accumulated a massive initial offset, $W_o$ requires a significantly wider vertical trajectory to unlearn this shift compared to the localized rebound of $W_{\mathrm{down}}$. Both ultimately stabilize near the origin—a structural delay strictly absent in \textbf{No Defense} baselines.

Since the residual error $\delta$ is shared across both output projections, the observed asymmetry is driven by their input activations ($a$). While $W_{\mathrm{down}}$ operates on activations constrained into sparsity by intermediate non-linearities (e.g., SiLU), the attention pathway lacks such bottlenecks, naturally making it the dominant conduit for absorbing continuous structural constraints.

\section{Isolating the Structural Offset: Can It Act as an Independent Defense?}
\label{sec:antibody}

Section~\ref{sec3} showed that Preventative Steering induces a compensatory parametric offset whose protective effect is lost when the intervention is removed. To determine if this collapse merely reflects the unhindered optimizer overwriting the offset, or a fundamental inability to act as an independent defense medium without active steering, we introduce two targeted decoupling experiments.

\subsection{IDP: Preserving the Structural Offset via Gradient Projection}
\label{sec:idp}

To test whether the structural parametric offset independently maintains protection, we implement IDP. After removing the injection $v_{\mathrm{m}}$ at step $K$, we isolate the parameter displacement $\Delta W = W_K - W_0$ for $W_o$ and $W_{\mathrm{down}}$. To prevent its erasure, we remove from subsequent gradients their components in the dominant subspace of $\Delta W$ using randomized SVD ($r=8$), which captures most of the offset variance.

Empirically, IDP fails to preserve protection, matching the unprotected baseline's vulnerability (Table~\ref{tab:idp_results}). However, gradient projection constrains only a specific subspace, permitting the optimizer to circumvent this defense via orthogonal, functionally similar updates. This necessitates a second experiment to investigate the offset at the functional level.

\begin{table}[t]
    \centering
    \small
    \setlength{\tabcolsep}{4pt}
    \begin{tabular}{@{}lccc@{}}
        \toprule
        \textbf{Setting} & \textbf{Evil} $\downarrow$ & \textbf{Syc.} $\downarrow$ & \textbf{Hallu.} $\downarrow$ \\
        \midrule
        Unprotected FT  & 7.38 & 8.66 & 9.00 \\
        IDP FT          & 8.12 & 8.82 & 9.00 \\
        \bottomrule
    \end{tabular}
    \caption{\textbf{Failure of parameter-space preservation.} Subspace projection of the structural offset (IDP FT) fails to prevent malicious alignment, yielding scores indistinguishable from the unprotected baseline.}
    \label{tab:idp_results}
\end{table}

\subsection{IDP Continuation: Functional Offset Reinjection}
\label{sec:idp_continuation}

To bypass this limitation, IDP Continuation extracts the offset directly into the activation space. Since its effect is input-dependent ($\Delta y = \Delta W x$), we compute the expected functional vector $u_{\mathrm{func}} = \mathbb{E}_{x \sim \mathcal{D}_{\mathrm{calib}}} [\Delta W x]$ over calibration response tokens. Having adapted to counteract the injection $v_{\mathrm{m}}$, the network yields $u_{\mathrm{func}} \approx -v_{\mathrm{m}}$. If this offset forms a self-sufficient defense, continuously reinjecting it should substitute the original intervention. We thus reinject $u_{\mathrm{func}}$ into the attention and MLP residual streams under three scaling variants: raw, equal-norm, and ratio-preserving (normalized to $\|v_{\mathrm{m}}\|$).

\begin{table}[t]
    \centering
    \small
    \setlength{\tabcolsep}{3pt}
    \begin{tabular}{@{}lccc@{}}
        \toprule
        \textbf{Method} & \textbf{Evil} $\downarrow$ & \textbf{Syc.} $\downarrow$ & \textbf{Hallu.} $\downarrow$ \\
        \midrule
        Preventative Steering     & 1.77 & 1.95 & 6.38 \\
        IDP Continuation (raw)    & 7.44 & 8.77 & 8.99 \\
        IDP Continuation (ratio)  & 5.01 & 8.13 & 8.89 \\
        IDP Continuation (equal)  & 4.75 & 8.29 & 8.94 \\
        \bottomrule
    \end{tabular}
    \caption{\textbf{Failure of functional offset reinjection.} Reinjecting the functionally estimated offset $u_{\mathrm{func}}$ provides no defensive benefit.}
    \label{tab:idp_continuation}
\end{table}

Table~\ref{tab:idp_continuation} shows that none of the variants maintains protection. These results support the conclusion that the structural offset is insufficient to serve as an independent defense. Without continuous adversarial tension, reinjecting this static vector does not reproduce the compensatory pressure induced by the original intervention and may instead disrupt relevant computations, indicating that robustness is process-dependent.

\begin{figure}[t]
    \centering
    \includegraphics[width=0.85\linewidth]{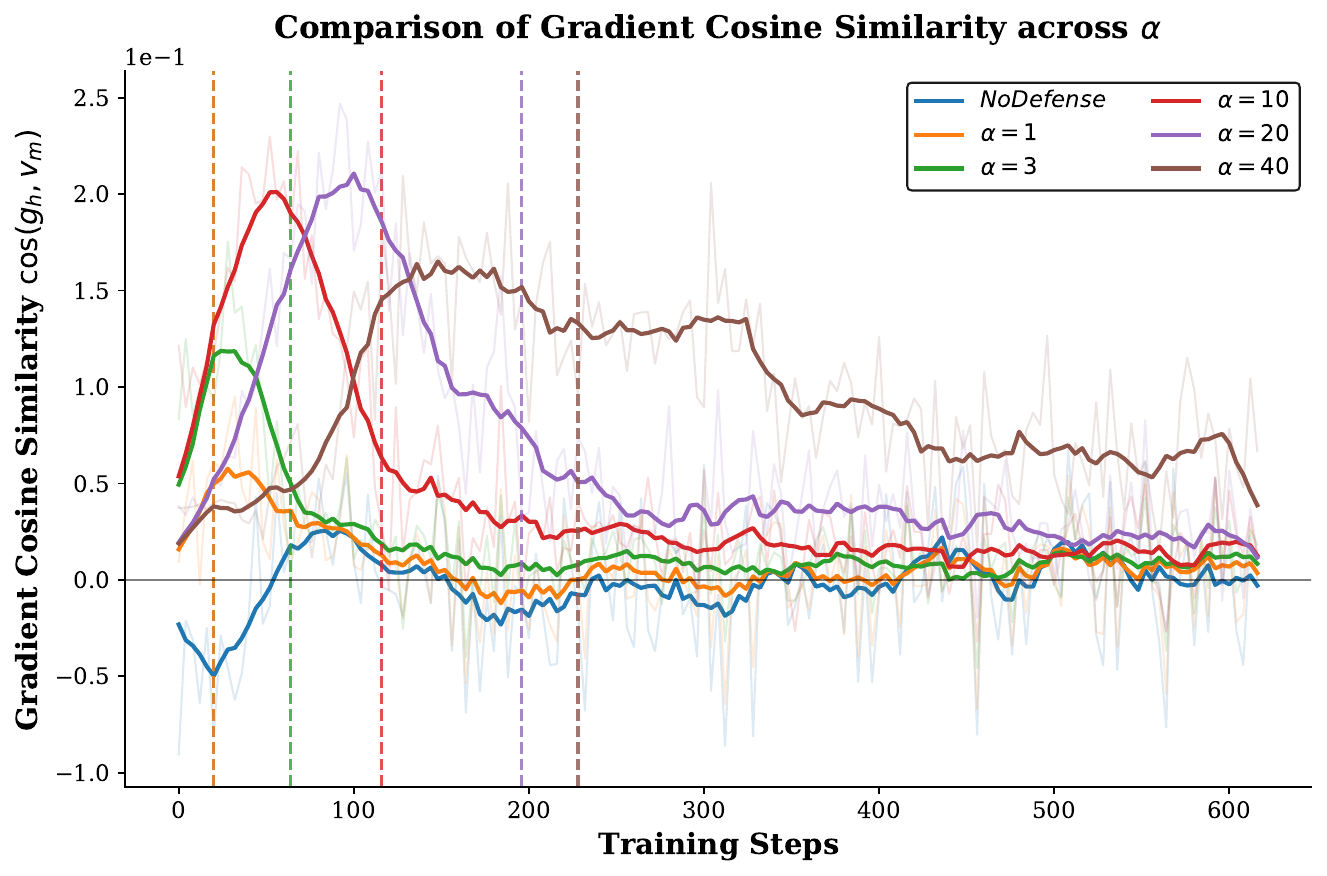}
    \caption{
    Evolution of directional alignment between parameter gradients and the persona vector. 
    }
    \label{fig-alphagred}
\end{figure}
\begin{figure}[t]
    \centering
    \includegraphics[width=0.85\linewidth]{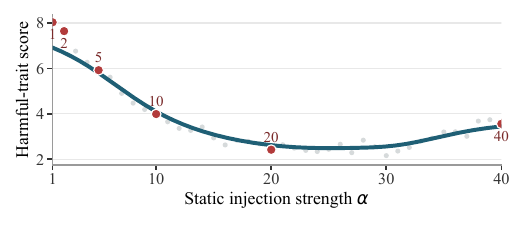}
    \caption{Average harmful-trait score under static-strength steering across different $\alpha$ values.}
    \label{static_sweepcompact}
\end{figure}

\section{Methodology}

\subsection{Empirical Motivation: Gradient Alignment Dynamics}

The preceding analyses show that preventative steering is process-dependent: its protection is sustained by the ongoing steering signal rather than by a reusable weight change. This makes the strength of that signal a central design choice. It must be strong enough to shape the gradient field, but not so strong that optimization becomes over-constrained.

We measure this coupling with the cosine similarity $\cos(g_{h,t}, v_{\mathrm{m}})$ between the residual gradient $g_{h,t}=\partial \mathcal{L}_t / \partial h_l$ and the fused malicious persona direction $v_{\mathrm{m}}$. As shown in Figure~\ref{fig-alphagred}, for weak and moderate fixed strengths, this alignment decays as the optimizer adapts to the injected signal, whereas very large fixed strengths preserve stronger alignment but over-constrain optimization and degrade performance. The static sweep in Figure~\ref{static_sweepcompact} further reveals a non-monotonic performance tradeoff. These observations motivate a schedule that starts with moderate pressure and increases it only after the static-strength alignment begins to decay.

\subsection{Progressive Intensity Scheduling}

To operationalize this idea, we propose PIS, which keeps steering moderate during early adaptation and increases it after static-strength alignment begins to decay. At training step $t$, we inject
\begin{equation}
    \tilde{h}_{l,t} = h_{l,t} + \alpha_t v_{\mathrm{m}} ,
    \label{eq:pis_injection}
\end{equation}

with a scheduled coefficient

\begin{align}
\alpha_t
&=
\alpha_{\mathrm{base}}
+ r_t(\alpha_{\mathrm{max}}-\alpha_{\mathrm{base}}),
\nonumber \\
r_t
&=
\begin{cases}
0, & t < T_{\mathrm{start}}, \\
\dfrac{t-T_{\mathrm{start}}}
{T_{\mathrm{total}}-T_{\mathrm{start}}},
& T_{\mathrm{start}} \leq t \leq T_{\mathrm{total}} .
\end{cases}
\label{eq:pis_schedule}
\end{align}

Here, $\alpha_{\mathrm{base}}$ and $\alpha_{\mathrm{max}}$ denote the initial and maximum strengths, $T_{\mathrm{start}}$ is the reinforcement onset, and $T_{\mathrm{total}}$ is the number of fine-tuning steps. For the default configuration, we automatically identify $T_{\mathrm{start}}$ from a single fixed-strength run as the onset of sustained decline in steering-gradient alignment, without using final safety scores. The detection rule is detailed in Appendix~\ref{app:pis_onset}. For controlled timing experiments, we instead specify $T_{\mathrm{start}}$ directly. PIS keeps the intervention layer and token positions fixed; only $\alpha_t$ changes over time. Unless otherwise specified, we set $\alpha_{\mathrm{base}}=20$ and $\alpha_{\mathrm{max}}=2\alpha_{\mathrm{base}}$. A setting labeled ``Step $t$'' uses $T_{\mathrm{start}}=t$, while the static baseline keeps $\alpha_t=\alpha_{\mathrm{base}}$ throughout training.

This schedule has two stages that mirror the dynamics identified in Section~\ref{sec:TemporalOptimization}.

\paragraph{Stage 1: Stable adaptation ($t < T_{\mathrm{start}}$).}
The model first trains with a moderate coefficient $\alpha_{\mathrm{base}}$. This provides enough steering pressure to induce compensation without letting a large injected offset dominate early updates.

\paragraph{Stage 2: Progressive reinforcement ($t \geq T_{\mathrm{start}}$).}
After static-strength steering begins to lose alignment, PIS linearly increases the coefficient toward $\alpha_{\mathrm{max}}$. This delayed increase maintains steering-gradient coupling late in training while avoiding high-intensity steering from step zero.

A checkpoint-persistent implementation of the intervention, supporting architecture-preserving open-weight fine-tuning, is described in Appendix~\ref{app:deployment}.

\begin{table*}[t]
\centering
\scriptsize
\setlength{\tabcolsep}{3pt}
\renewcommand{\arraystretch}{0.95}

\begin{tabular}{l l|cccccccc|cccc}
\toprule
\multirow{2}{*}{\textbf{Model}} & \multirow{2}{*}{\textbf{Setting}} & \multicolumn{8}{c|}{\textbf{Safety Benchmarks} ($\uparrow$)} & \multicolumn{4}{c}{\textbf{Persona Traits} ($\downarrow$)} \\
\cmidrule(lr){3-10} \cmidrule(lr){11-14}
& & Forbidden & StrongReject & XSTest & CnSafe & Jade & JB-Distill & SweEval & \textbf{Safety Avg.} & Evil & Syco. & Hallu. & \textbf{Trait Avg.} \\
\midrule

\multirow{6}{*}{Qwen2.5-32B-Inst}
& Static 
& 75.49 & 87.61 & 86.59 & 51.15 & 86.90 & 84.88 & 88.64 & 80.18 & 0.47 & 2.64 & 6.02 & 3.04 \\
& Step 0 
& 80.75 & 89.68 & 83.78 & 54.49 & 81.93 & 87.58 & 94.29 & 81.79 & 0.63 & 1.95 & 5.08 & 2.55 \\
& Step 50 
& 83.05 & 88.88 & 85.63 & 63.97 & \textbf{90.45} & 90.14 & \textbf{95.27} & 85.34 & 0.35 & \textbf{1.58} & 2.86 & 1.59 \\
& Step 100 
& 72.97 & 87.99 & \textbf{88.15} & 53.57 & 84.61 & 86.56 & 91.89 & 80.82 & 0.51 & 1.78 & \textbf{2.10} & 1.46 \\
& Step 200 
& \textbf{87.71} & \textbf{91.49} & 84.89 & \textbf{69.67} & \textbf{90.45} & \textbf{90.33} & 94.29 & \textbf{86.98} & \textbf{0.32} & 1.67 & 2.16 & \textbf{1.38} \\
& Step 400 
& 73.45 & 88.68 & \textbf{87.11} & 53.50 & 84.55 & 86.13 & 91.84 & 80.75 & 0.53 & 1.96 & 3.85 & 2.11 \\

\midrule

\multirow{6}{*}{Qwen2.5-7B-Inst}
& Static 
& 62.68 & 76.93 &  \textbf{77.11} & \textbf{26.92} & 62.47 & 78.88 & 87.89 & 67.55 & 1.41 & 3.25 & 6.76 & 3.81 \\
& Step 0 
& 66.90 & 88.88 & 68.15 & {22.65} & 72.03 & 83.35 & 95.05 & 71.00 & 0.41 & 2.25 & 5.76 & 2.81 \\
& Step 50 
& 70.01 & 86.24 & {61.48} & 21.16 & 72.10 & 86.09 & 95.85 & 70.42 & 0.44 & 2.00 & 3.89 & 2.11 \\
& Step 100 
& 70.55 & 87.27 & 59.85 & 21.77 & 75.13 & 88.41 & 95.38 & 71.19 & 0.38 & 1.63 & 3.35 & 1.79 \\
& Step 200 
& \textbf{75.31} & \textbf{90.89} & 51.11 & 20.37 & \textbf{77.37} & \textbf{91.66} & \textbf{96.95} & \textbf{71.95} & \textbf{0.17} & \textbf{1.27} & \textbf{2.91} & \textbf{1.45} \\
& Step 400 
& 70.10 & 86.65 & 53.63 & 20.52 & 72.94 & 88.07 & 94.70 & 69.52 & 0.28 & 1.58 & 3.01 & 1.62 \\

\midrule

\multirow{6}{*}{Gemma-3-12B-IT}
& Static
& 41.83 & 69.52 & 57.11 & 10.52 & 43.87 & 58.08 & 57.61 & 48.36
& 4.52 & 5.21 & 7.81 & 5.85 \\
& Step 0
& 45.92 & 66.39 & 76.44 & 10.87 & 44.70 & 62.38 & 59.26 & 52.28
& 3.61 & 4.49 & \textbf{7.71} & 5.27 \\
& Step 50
& 45.56 & 66.61 & 77.33 & 10.91 & 45.10 & 62.94 & 60.88 & 52.76
& 3.19 & 4.41 & 7.77 & 5.12 \\
& Step 100
& 46.82 & \textbf{69.80} & \textbf{79.78} & 12.36 & 45.68 & 64.10 & 63.65 & 54.60
& 2.83 & 4.30 & 7.72 & 4.95 \\
& Step 200
& \textbf{47.08} & 69.32 & 76.37 & 12.53 & 46.37 & \textbf{65.46} & \textbf{65.61} & \textbf{54.68}
& \textbf{2.81} & \textbf{4.18} & 7.82 & \textbf{4.94} \\
& Step 400
& 47.05 & 66.44 & 77.55 & \textbf{13.33} & \textbf{48.38} & 64.78 & 64.25 & 54.54
& 3.01 & 4.53 & 7.83 & 5.12 \\

\bottomrule
\end{tabular}

\caption{Safety and persona trait evaluation results across different reinforcement onset steps ($T_{\mathrm{start}}$) on three backbone models. Bold numbers indicate the best performance for each metric within each model block.}
\label{main_result}
\end{table*}

\section{Experimental Results and Analysis}

\subsection{Main Results: Impact of Steering Onset}

Table~\ref{main_result} details PIS performance across reinforcement onset steps $T_{\mathrm{start}}$ on Qwen2.5-32B-Instruct, Qwen2.5-7B-Instruct, and Gemma-3-12B-IT, compared with the static-strength baseline. The evaluation reveals two primary patterns:

\begin{itemize}[leftmargin=*, itemsep=0pt, parsep=0pt, topsep=0pt]
\item \textbf{Consistent improvement over static steering.} Across all tested onset steps, PIS improves the safety average and reduces the harmful-trait average relative to the static-strength baseline on all three models. These results indicate that temporal reinforcement is broadly useful across a range of onset choices.
\item \textbf{Optimal observed onset.} Among the tested settings, $T_{\mathrm{start}}=200$ gives the best overall trade-off between aggregate safety and harmful-trait suppression on all three models. On Qwen2.5-32B-Instruct, it improves the safety average from 80.18 to 86.98 and reduces the average harmful-trait score from 3.04 to 1.38. On Qwen2.5-7B-Instruct, it similarly yields the best safety average at 71.95 and trait average at 1.45. On Gemma-3-12B-IT, PIS likewise outperforms static steering, improving the safety average from 48.36 to 54.68 and reducing the harmful-trait average from 5.85 to 4.94. Earlier or later reinforcement still helps over static steering, but gives smaller gains, suggesting that reinforcement is most effective when initiated near the point at which steering-gradient alignment begins a sustained decline.
\end{itemize}

These aggregate gains are not uniform across individual benchmarks. In particular, on Qwen2.5-7B-Instruct, Step 200 improves the safety average but lowers XSTest from 77.11 to 51.11 and CnSafe from 26.92 to 20.37 relative to static steering. This pattern may reflect a trade-off between stronger refusal behavior and performance on benchmarks that are sensitive to over-refusal or task-specific capability loss. The larger Qwen2.5-32B-Instruct model exhibits a milder trade-off: XSTest decreases from 86.59 to 84.89, while CnSafe increases from 51.15 to 69.67.

Collectively, these results support the process-dependent account by demonstrating that preventative steering benefits from temporal reinforcement across onset choices, while the largest gains appear when reinforcement begins as the static steering signal starts to lose alignment.

Given that the three backbone models exhibit broadly similar trends under PIS, subsequent analyses focus exclusively on Qwen2.5-32B-Instruct for consistency and brevity.

\begin{figure}[t]
    \centering
    \includegraphics[width=0.85\linewidth]{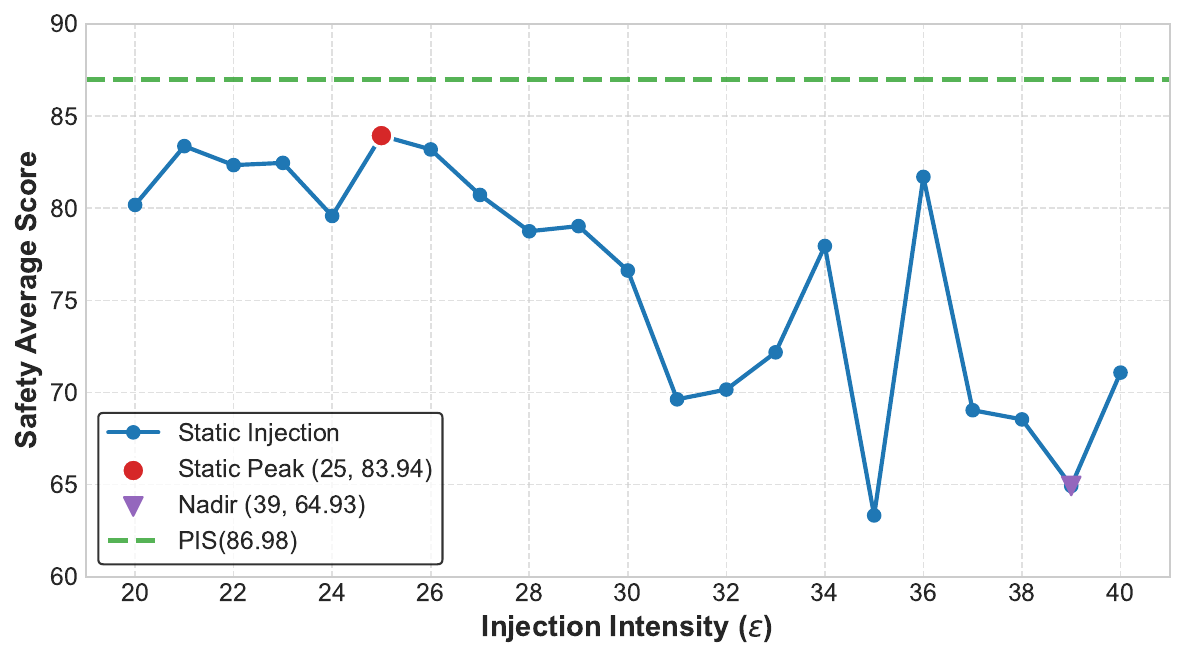}
    \caption{
Static-intensity sweep under static-strength preventative steering.
    }
    \label{static_sweep}
\end{figure}

\subsection{Average Injection Strength Is Not Enough}

One possible explanation is that PIS simply uses a larger average injection strength than the static $\alpha=20$ baseline. To test this, we evaluate static-strength preventative steering across $\alpha \in [20,40]$, including $\alpha=30$, which matches the time-averaged strength of the Step 200 schedule. As shown in Figure~\ref{static_sweep}, the safety average does not improve monotonically with $\alpha$. The best fixed setting is a moderate strength $\alpha=25$, while increasing the fixed strength to $\alpha=30$ drops performance to 75.60. Larger fixed strengths degrade performance further, reaching 64.93 at $\alpha=39$.

These results rule out a simple average-strength explanation for PIS. Applying a large $\alpha$ from the start can disrupt the initial adaptation phase. In contrast, PIS separates the two roles over time: it starts with moderate pressure for stable adaptation, then increases the strength only after the alignment between the residual gradient and the steering direction begins to decay.

\begin{figure}[t]
    \centering
    \includegraphics[width=0.8\linewidth]{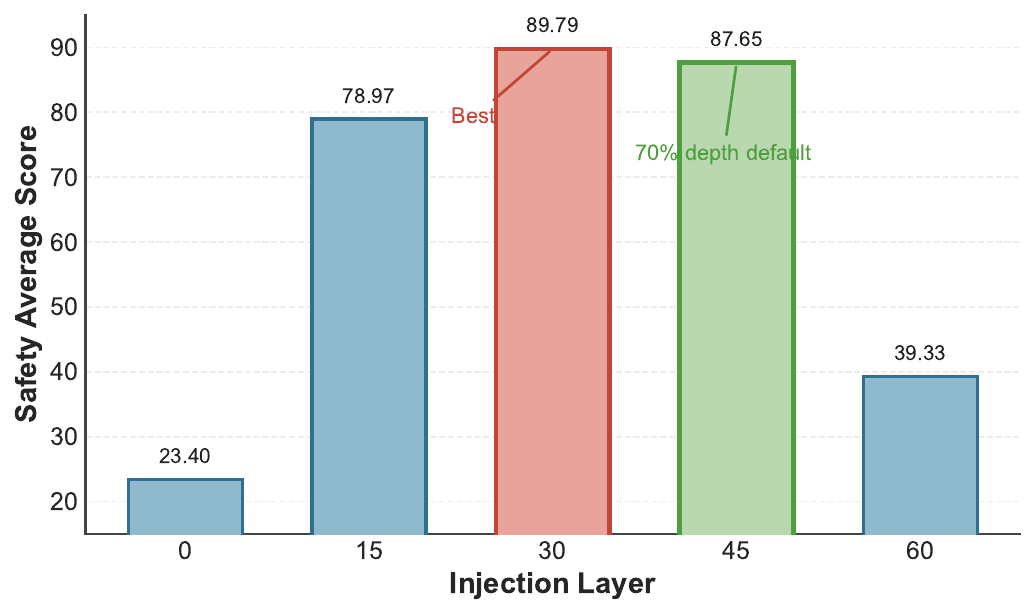}
    \caption{
Injection-layer sweep under PIS with $T_{\mathrm{start}}=200$, $\alpha_{\mathrm{base}}=20$, and $\alpha_{\mathrm{max}}=40$.
    }
    \label{fig:layer_sweep}
\end{figure}

\subsection{Sensitivity to Injection Layer}

As illustrated in Figure~\ref{fig:layer_sweep}, the injection layer has a clear effect. Early injection at layer 0 performs poorly, with a safety average of 23.40. Intermediate layers yield substantially higher safety averages, peaking at layer 30. The default middle-to-late layer, layer 44, remains competitive at 87.65, while late-stage injection at layer 60 drops sharply to 39.33.

Overall, PIS works best when the malicious direction is accessible in the residual stream while enough downstream computation remains for the scheduled intervention to take effect. Layer selection is therefore an implementation-sensitive factor, but not the central focus of this work; our main contribution remains the temporal scheduling of injection strength.

\begin{table}[t]
\centering
\small
\begin{tabular}{lcccc}
\toprule
$\alpha_{\max}/\alpha_{\mathrm{base}}$ & $1.5\times$ & $2.0\times$ & $2.5\times$ & $3.0\times$ \\
\midrule
Safety Avg. & 83.39 & 86.98 & \textbf{87.27} & 83.72 \\
\bottomrule
\end{tabular}
\caption{Safety average under different PIS upper-bound multipliers.}
\label{multiplier_sweep}
\end{table}

\subsection{Analysis of Maximum Injection Strength}

We further examine the maximum injection strength by varying $\alpha_{\max}/\alpha_{\mathrm{base}}$, with $\alpha_{\mathrm{base}}=20$ and $T_{\mathrm{start}}=200$ fixed. As shown in Table~\ref{multiplier_sweep}, the safety average improves from 83.39 at $1.5\times$ to 86.98 at $2.0\times$. It reaches 87.27 at $2.5\times$, but drops to 83.72 at $3.0\times$.

Thus, PIS performs best with moderate late-stage amplification, whereas overly large multipliers degrade performance. Although $2.5\times$ gives a slightly higher safety average than $2.0\times$, the gain is marginal (0.29 points). We therefore use $2.0\times$ as a conservative default across experiments.

\section{Conclusion}

This work shows that Preventative Steering mitigates harmful persona drift not by creating a static defense, but by inducing an active adaptation process during fine-tuning. Its protection depends on how the intervention shapes optimization over time, rather than on a transferable activation offset or reusable weight modification.

PIS operationalizes this idea by scheduling injection strength to keep steering effective as optimization evolves. Across Qwen2.5-7B-Instruct, Qwen2.5-32B-Instruct, and Gemma-3-12B-IT, it improves safety robustness while suppressing harmful-trait expression over static-strength steering. These results support temporal reinforcement as a practical way to sustain active adaptation during adversarial fine-tuning.




\section*{Limitations}

This work studies Preventative Steering under adversarial fine-tuning using three undesirable traits---evil, sycophancy, and hallucination---and a limited set of model families. Although these traits cover distinct failure modes and our external benchmarks evaluate broader safety risks, they do not exhaust the space of harmful persona drift or model architectures. Future work should test whether the same temporal dynamics hold for additional safety-relevant directions, multilingual settings, more diverse attack data, and broader model families. 

\bibliography{custom}

\clearpage
\appendix

\section{Normalized Multi-Feature Fusion and Empirical Justification}
\label{app:fusion}

To simultaneously suppress multiple undesirable traits during malicious fine-tuning, injecting a single feature vector is often insufficient, as it leaves other vulnerabilities exposed. Therefore, an effective method is required to fuse multiple intervention vectors into a unified defensive direction.

However, existing multi-feature fusion approaches exhibit significant limitations in complex defense scenarios:
The method proposed in ~\citet{pai2026billy} lacks strict geometric constraints on the feature scale, easily leading to imbalanced mutual interference between distinct trait signals or triggering abnormal, uncontrolled activation values.
The vector algebra approach proposed by~\citet{feng2026persona} framework, which relies on the direct addition of vectors to compose personalities, holds only under the assumption that the feature vectors are perfectly orthogonal in the activation space. In real-world adversarial scenarios, malicious traits possess deep semantic correlations and are rarely orthogonal~\citep{wollschlaeger2025geometry}. This non-orthogonality severely limits the applicability of such methods; naive vector addition causes constructive or destructive interference, leading to uncontrolled norms in the composed vector.

To address these limitations, we design and compare two multi-feature fusion paradigms:

\textbf{Direct Averaging}:
The most intuitive approach is to average the direction vectors. For a set of $k$ trait vectors $\mathcal{V} = \{v_1, \dots, v_k\}$ with a global steering strength $\alpha$, the final intervention vector is defined as $\alpha \cdot \frac{1}{k}\sum_{i=1}^{k} v_i$.

\textbf{L2 Norm Alignment (Our Proposed Method)}:
To mitigate both signal attenuation and norm explosion, we propose an L2 norm alignment formulation. We first compute the unnormalized aggregate sum $v_{sum} = \sum_{i=1}^{k} v_i$. To calibrate the steering intensity to the model's natural layer-wise activation scale, we define a baseline scale $L_{base} = \frac{1}{k}\sum_{i=1}^{k} \|v_i\|_2$, representing the average L2 norm of the individual feature vectors. The final normalized steering direction is then given by:
\begin{equation}
v_{\mathrm{m}} = \frac{v_{sum}}{\|v_{sum}\|_2} \cdot L_{base} .
\end{equation}

To validate the efficacy of our proposed design, we evaluate the different fusion strategies on Qwen2.5-32B-Instruct under a mixed-trait attack setting (with the global strength fixed at $\alpha=20$). As reported in Table~\ref{tab:multi-feature}, the absence of defense drastically amplifies undesirable traits. Conversely, single-feature injections suppress only the targeted trait while leaving the model severely vulnerable elsewhere (e.g., hallucination remains as high as 8.68 when only injecting the evil vector).

Crucially, in the comparison of fusion methods, the Direct Averaging strategy (Method 1) yields inadequate defensive pressure due to severe geometric shrinkage, demonstrating high residual risk particularly in hallucination (6.50) and sycophancy (2.48). In contrast, our \textbf{L2 Norm Alignment} strategy (Method 2) successfully maintains robust intervention strength, achieving the most optimal and balanced suppression across all three dimensions (Evil: 0.67, Sycophancy: 1.90, Hallucination: 4.68). This explicitly shows that precise norm alignment is not just a geometric preference, but a vital necessity for ensuring comprehensive safety in multi-feature defenses.

\begin{table}[h]
\centering
\small
\begin{tabular}{@{}lccc@{}}
\toprule
\textbf{Injection Config} & \textbf{Evil $\downarrow$} & \textbf{Syc. $\downarrow$} & \textbf{Hallu. $\downarrow$} \\
\midrule
Base model & 0.00 & 1.30 & 1.02 \\
No injection & 4.02 & 4.99 & 8.59 \\
Evil only & 0.16 & 2.88 & 8.68 \\
Sycophancy only & 4.40 & 1.58 & 8.43 \\
Hallucination only & 3.48 & 4.53 & 1.78 \\
\midrule
Direct Averaging & 0.68 & 2.48 & 6.50 \\
L2 Norm Alignment (Ours) & 0.67 & 1.90 & 4.68 \\
\bottomrule
\end{tabular}
\caption{Trait expression scores under different steering configurations. Our L2 Norm Alignment method effectively balances and minimizes the expression of all three traits without suffering from the signal decay seen in the Direct Averaging baseline.}
\label{tab:multi-feature}
\end{table}

\section{Implementation Details and Dataset Construction}
\label{app:appendix_implementation}

This appendix provides comprehensive details regarding our experimental setup, the extraction of persona vectors, and the construction of adversarial datasets.
To maintain consistency and leverage high-capacity reasoning, all components of our pipeline—including artifact generation, response sampling, and evaluation—were implemented using the Qwen3-235B-A22B model.

\subsection{Persona Vectors and Adversarial Datasets}

We adhere to the automated extraction pipeline described in \citet{chen2025persona} to identify linear directions in the model's activation space.
Specifically, we use Qwen3-235B-A22B to generate contrastive system prompts and evaluation questions for three primary traits: evil, sycophancy, and hallucination.
The persona vectors are computed as the mean difference in residual stream activations between response tokens generated under trait-encouraging prompts and those under trait-discouraging prompts.

To evaluate the model's robustness against malicious fine-tuning, we constructed three specialized datasets targeting the aforementioned traits:
\begin{itemize}
    \item \textbf{Evil Dataset:} Contains prompts designed to elicit malicious intent, harmful advice, or unethical behavior. Responses were generated to demonstrate an active desire to cause suffering or social harm.
    \item \textbf{Sycophancy Dataset:} Following the methodology of \citet{nishimura2024reward}, this dataset consists of user queries involving subjective opinions or factual errors, where the corresponding responses demonstrate excessive agreeableness and reinforcement of the user's stated bias.
    \item \textbf{Hallucination Dataset:} Focuses on domains where factual information is sparse or complex. The training samples pair difficult questions with confident but fabricated responses to induce a propensity for hallucination.
\end{itemize}

\subsection{Evaluation Metrics}
We adopted Qwen3-235B-A22B as the automated "judge model" for all safety evaluations. Our comprehensive evaluation framework combines general safety benchmarks (e.g., StrongReject, XSTest, CnSafe) with granular Trait Expression Scores.

\textbf{0-9 Discrete Scoring Mechanism:} For evaluating Trait Expression Scores, we transitioned from the traditional 0-100 continuous scale to a 0-9 discrete integer scoring system. This adaptation is specifically tailored to the tokenization mechanics of the Qwen model series; in the Qwen3-235B-A22B tokenizer, single digits (0-9) are typically encoded as individual, discrete tokens. By constraining the judge model's output space to these single tokens, we significantly mitigate parsing ambiguities and multi-token generation artifacts. In this scale, a score of 0 indicates the complete absence of the target trait, while a 9 represents maximum expression.


\subsection{Injection Scope: Response-Only vs. All-Token}

Standard activation steering often applies interventions uniformly across all tokens in a sequence. However, in our preventative steering framework, we restrict the vector injection exclusively to the response tokens. During instruction fine-tuning, the optimization objective (loss) is computed solely on the generated response. Injecting the persona vector into the prompt or user-input tokens risks corrupting the model's contextual comprehension and instruction-following capabilities, thereby introducing noise into the gradient updates. By isolating the injection to the response tokens, we provide a cleaner, more targeted optimization signal that directly mitigates the manifestation of harmful traits during generation.

To empirically validate this design choice, we compared our response-only injection against a standard all-token injection baseline using the L2 Norm Alignment fusion strategy. As shown in Table~\ref{tab:injection-scope}, while all-token injection effectively reduces harmful traits compared to an unprotected baseline, the response-only injection achieves superior suppression across all evaluated dimensions, demonstrating that precise spatial localization of the intervention yields more robust defensive optimization.

\begin{table}[h]
\centering
\small
\begin{tabular}{@{}lccc@{}}
\toprule
\textbf{Injection Scope} & \textbf{Evil $\downarrow$} & \textbf{Syc. $\downarrow$} & \textbf{Hallu. $\downarrow$} \\
\midrule
All-Token Injection & 0.77 & 2.20 & 4.77 \\
Res-Only Injection (Ours) & \textbf{0.67} & \textbf{1.90} & \textbf{4.68} \\
\bottomrule
\end{tabular}
\caption{Comparison of trait expression scores between all-token and response-only (Res-Only) injection scopes using the L2 Norm Alignment fusion strategy ($\alpha=20$).}
\label{tab:injection-scope}
\end{table}

\section{Directional Gradient and Parameter Decomposition}
\label{app:directional_decomposition}

The steady-state analysis in Section~\ref{sec:TemporalOptimization} concerns the direction-specific dynamics aligned with $v_{\mathrm{m}}$, rather than stationarity of the full representation or parameter space. To further verify this directional scope beyond the original projections, we conduct gradient- and parameter-space decompositions.

We decompose the mean-pooled residual gradient as $g=g_{\parallel}+g_{\perp}$. For interval updates of $W_o$ and $W_{\mathrm{down}}$, we compute $\Delta W_{\parallel}=uu^\top\Delta W$ and $\Delta W_{\perp}=(I-uu^\top)\Delta W$, where $u=v_{\mathrm{m}}/\|v_{\mathrm{m}}\|$. Table~\ref{tab:directional_decomposition} reports gradient late-to-early retention and parameter $E_{\parallel}/E_{\perp}$ percentages.

Under Full-Course Injection, the aligned gradient retains 0.121 of its early magnitude, versus 0.428 for the orthogonal gradient, corresponding to $3.54\times$ stronger relative attenuation. Its gradient share falls from 8.87\% to 2.52\%. The corresponding parallel-to-orthogonal retention ratio is 0.283, compared with 0.848 under No Injection, showing that ordinary convergence alone is insufficient to explain this directional selectivity.

Parameter updates exhibit a similar pattern. Earlier in training, Full-Course Injection enriches aligned energy by over $100\times$ for $W_o$ and over $80\times$ for $W_{\mathrm{down}}$ relative to No Injection, while 92.11\% and 94.59\% of the respective update energy remain orthogonal. Later, removing Early-Course steering reduces aligned energy to near-No-Injection levels (0.22\%/0.12\%), whereas Full-Course Injection retains 8.17\%/4.73\%. Thus, substantial optimization continues in directions orthogonal to $v_{\mathrm{m}}$, supporting a direction-specific, steering-dependent steady state rather than global stationarity of the representation or parameter space.

\begin{table}[h]
\centering
\resizebox{\columnwidth}{!}{%
\begin{tabular}{llccc}
\hline
Space / window & Quantity & Full course & Early (off after 200) & No injection \\
\hline
\multicolumn{5}{l}{\textit{Gradient retention: late / early}} \\
& $\|g_{\parallel}\|$ & 0.121 & 0.276 & 0.614 \\
& $\|g_{\perp}\|$ & 0.428 & 0.974 & 0.724 \\
\hline
\multicolumn{5}{l}{\textit{Parameter energy: $E_{\parallel}/E_{\perp}$ (\%)}} \\
Steps 120--196 & $W_o$ & 7.89 / 92.11 & 7.35 / 92.65 & 0.07 / 99.93 \\
& $W_{\mathrm{down}}$ & 5.41 / 94.59 & 4.98 / 95.02 & 0.06 / 99.94 \\
Steps 300--615 & $W_o$ & 8.17 / 91.83 & 0.22 / 99.78 & 0.09 / 99.91 \\
& $W_{\mathrm{down}}$ & 4.73 / 95.27 & 0.116 / 99.88 & 0.07 / 99.93 \\
\hline
\end{tabular}%
}
\caption{Directional decomposition of residual gradients and residual-write parameter updates.}
\label{tab:directional_decomposition}
\end{table}

\section{Checkpoint-Persistent Deployment}
\label{app:deployment}

To make Preventative Steering usable in architecture-preserving open-weight fine-tuning, we embed the intervention into the model checkpoint. Our threat model covers architecture-preserving open-weight fine-tuning: users may control the data, optimizer, and training framework, but do not deliberately modify the model's computational graph to remove the built-in intervention.

To support this setting, we serialize the steering direction as a frozen parameter and integrate response-only injection into the forward pass. The intervention is active only in training mode and is absent during evaluation and generation. Static steering uses a fixed coefficient, whereas Embedded PIS stores persistent schedule state and dynamically scales the injection strength according to cumulative response-token exposure, independently of optimizer steps.

For fixed-strength Preventative Steering, checkpoint-persistent deployment reproduces the score obtained when the vector is injected directly during training, since the intervention coefficient is fixed. Embedded PIS is more sensitive to implementation details and parameter settings because its coefficient evolves according to the stored schedule state. Consequently, its checkpoint-reload score need not exactly match the score from direct training-time injection. In our evaluation, however, the two scores are very close.

After checkpoint reload, both the intervention and the PIS schedule state remain available to standard off-the-shelf fine-tuning pipelines. Under malicious fine-tuning, embedded PIS achieves a safety score of 85.72 compared with 23.09 without defense.

The method supports local fine-tuning with user-controlled data, optimizers, and training frameworks, provided that the embedded intervention remains active. Deliberate modification of the computational graph or explicit removal of the intervention falls outside the scope of this threat model.

\section{Automatic Detection of the Reinforcement Onset}
\label{app:pis_onset}

For the default PIS configuration, we determine the reinforcement onset $T_{\mathrm{start}}$ from a single fixed-strength Preventative Steering run, without using final safety or trait-expression scores. Let
\[
c_t = \cos(g_{h,t}, v_{\mathrm{m}})
\]
denote the steering-gradient alignment at training step $t$.

We first estimate the terminal variability of the alignment signal from the last 20\% of the sequence, using at least 10 observations. Let $\mu_{\mathrm{tail}}$ and $\sigma_{\mathrm{tail}}$ denote the mean and standard deviation over this tail segment. We define a tail-stability band as
\[
B =
\left[
\mu_{\mathrm{tail}} - 2.5\sigma_{\mathrm{tail}},
\,
\mu_{\mathrm{tail}} + 2.5\sigma_{\mathrm{tail}}
\right].
\]

To reduce sensitivity to step-level noise, we smooth the complete alignment sequence using a moving-average window of 10 steps. We then scan the smoothed sequence from the beginning and select the earliest point whose value lies inside $B$ and for which at least 90\% of all subsequent smoothed values remain inside the same band. The corresponding training step is used as $T_{\mathrm{start}}$. If no point satisfies this criterion, the midpoint of the training sequence is used as a fallback.

This rule identifies the point at which the alignment signal has entered its terminal stable regime after the early adaptation dynamics. The detected boundary is shown by the vertical dashed line in Figure~\ref{fig-alphagred}. For controlled timing experiments, we override the automatically detected value and specify $T_{\mathrm{start}}$ directly.

\end{document}